\documentclass[sigconf,nonacm]{acmart}
\usepackage{booktabs} 

\usepackage{cuted}
\usepackage{capt-of}

\usepackage[ruled]{algorithm2e} 
\usepackage{multirow} 

\SetAlFnt{\small}
\SetAlCapFnt{\small}
\SetAlCapNameFnt{\small}
\SetAlCapHSkip{0pt}

\def\papername{X-UniMotion}

\begin{document}
\title{\papername: Animating Human Images with Expressive, Unified and Identity-Agnostic Motion Latents}


\author{Guoxian Song}
\affiliation{%
 \institution{ByteDance}
 \country{USA}
 }
\email{guoxiansong@bytedance.com}

\author{Hongyi Xu}
\affiliation{%
 \institution{ByteDance}
 \country{USA}
 }
\email{hongyixu@bytedance.com}

\author{Xiaochen Zhao}
\affiliation{%
 \institution{ByteDance}
 \country{USA}
 }
\email{xiaochen.zhao@bytedance.com}

\author{You Xie}
\affiliation{%
 \institution{ByteDance}
 \country{USA}
 }
\email{you.xie@bytedance.com}

\author{Tianpei Gu}
\affiliation{%
 \institution{ByteDance}
 \country{USA}
 }
\email{tianpei.gu@bytedance.com}

\author{Zenan Li}
\affiliation{%
 \institution{ByteDance}
 \country{USA}
 }
\email{zenan.li@bytedance.com}

\author{Chenxu Zhang}
\affiliation{%
 \institution{ByteDance}
 \country{USA}
 }
\email{chenxuzhang@bytedance.com}

\author{Linjie Luo}
\affiliation{%
 \institution{ByteDance}
 \country{USA}
 }
\email{linjie.luo@bytedance.com}

\begin{strip}
	\centering
	\includegraphics[width=\linewidth]{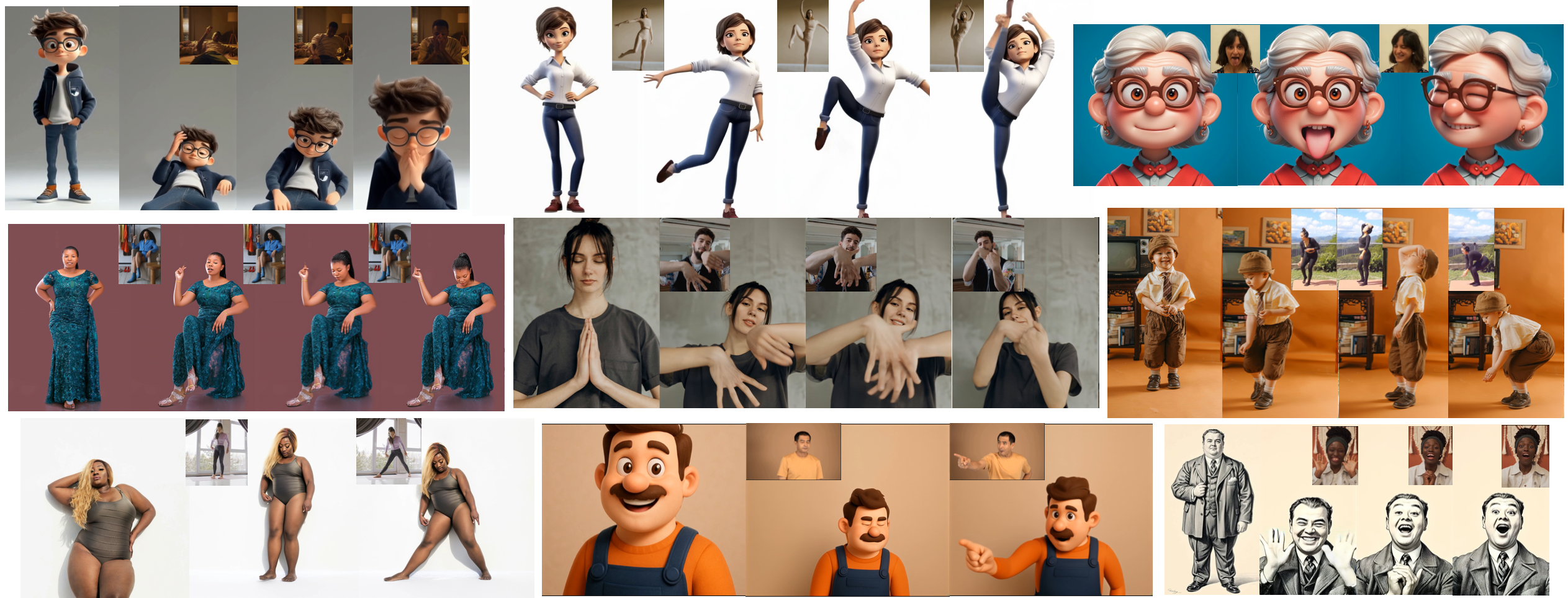}
	\centering
    \vspace{-6mm}
	\captionof{figure}{\papername~ introduces an expressive, unified and identity-agnostic latent human motion representation at full-body scale—including facial expressions and hand gestures—enabling fine-grained, depth-aware video reenactment across characters with diverse diverse identities, poses and spatial configurations.   
 }  
\label{fig:teaser}
 \vspace{-3mm}
\end{strip}

\begin{abstract}
We present \papername, a unified and expressive implicit latent representation for whole-body human motion, encompassing facial expressions, body poses, and hand gestures. Unlike prior motion transfer methods that rely on explicit skeletal poses and heuristic cross-identity adjustments, our approach encodes multi-granular human motion directly from a single image into a compact set of four disentangled latent tokens—one each for facial expression and body pose, and one per hand. These motion latents are both highly expressive and identity-agnostic, enabling high-fidelity, detailed cross-identity motion transfer across subjects with distinct identities, poses and spatial configurations.
To achieve this, we introduce a self-supervised, end-to-end training framework that jointly learns the motion encoder and latent representation alongside a DiT-based video generative model, trained on large-scale video datasets spanning diverse human motions. 
Motion-identity disentanglement is enforced via 2D spatial and color augmentations, as well as synthetic 3D renderings of cross-identity subject pairs under shared poses. We further guide the learning of motion tokens using auxiliary decoders to promote fine-grained, semantically aligned, and depth-aware motion embeddings.
Extensive experiments demonstrate that \papername~outperforms state-of-the-art methods, producing highly expressive animations with superior motion expressiveness and identity preservation.



\end{abstract}


%
%


\maketitle

\section{Introduction}

Animating a single human image using motions derived from a driving video of a different subject is a compelling yet challenging task, with wide-ranging applications in video conferencing, avatar animation, and digital humans. The core challenge lies in faithfully modeling and transferring whole-body human motion—including facial expressions, body poses, and hand gestures—while preserving the identity and appearance of the target subject.

Most existing motion transfer methods rely on explicitly extracted 2D or 3D pose representations, such as skeletal keypoints obtained from detectors like MMPose~\cite{mmpose2020} and Sapiens~\cite{khirodkar2024sapiens}. However, these representations are often limited in expressiveness, particularly in capturing fine-grained motions or dynamic cues, and suffer from inherent 2D depth ambiguities (e.g., finger or limb crossings). Furthermore, the extracted motion signals are frequently entangled with identity-specific attributes and  require heuristic adjustments or manual calibration for cross-identity transfer. These limitations hinder the generation of high-fidelity, identity-consistent animations, especially in cross-subject scenarios involving significant differences in body structure, pose articulation and spatial coverage.

In this work, we introduce \papername, a unified and expressive implicit latent representation for full-body human motion. Our key insight is to move away from explicit pose inputs and instead encode identity-agnostic motion implicitly from a single image. We represent motion as a set of four disentangled latent tokens: one each for facial expression, body pose, and each hand. Despite their minimal form, these latents capture rich, multi-scale motion signals—including subtle facial expressions, intricate finger gestures, complex body articulation, and even dynamic motion cues such as motion blur. 
Moreover, our representation is inherently free from the 2D depth ambiguities found in skeletal pose maps, enabling more accurate modeling of self-occlusions and overlapping limbs. 
Critically, our motion tokens are trained to be disentangled from identity-related attributes such as face shape, body proportions, clothing, and hairstyle. Extracted directly from raw images without reliance on external pose detectors, the latents support lossless motion derivation and are robust under challenging occlusions and lightings. This also enables an end-to-end learning paradigm, wherein the motion encoder is trained jointly with a motion-conditioned video generator. \papername~enables highly expressive, identity-preserving, and robust human image animation, even in challenging cross-identity scenarios involving vastly differences in identity and pose attributes.

To enable this, we develop a self-supervised, end-to-end training framework in which a ViT-based~\cite{dosovitskiy2020image} image encoder extracts motion latents from video frames, while a DiT model~\cite{peebles2023scalableDiT} is trained to generate future frames from noise, conditioned on these motion embeddings.  This setup enables us to fully leverage the generative capacity of the DiT model—pretrained on large-scale video datasets—to elicit fine-grained, expressive motion representations and produce high-fidelity human animations driven by the encoded motion tokens.
Motion-identity disentanglement is enforced through 2D motion-invariant augmentations (e.g., spatial transformations and color jittering) applied to the driving frames, as well as through synthetic 3D renderings of cross-identity subject pairs with shared poses. To capture fine-grained facial expressions and hand gestures, we supplement the global motion descriptor with three localized tokens extracted from face and hand crops.
To retarget disentangled motion to a reference subject, we use a ViT decoder that fuses motion latents with reference image patches, reasoning over spatial alignment consistent with the reference subject’s identity and structure. To further enhance the semantic alignment and granularity of the motion tokens, we introduce auxiliary supervision via a set of lightweight dual decoder heads: a GAN-based decoder for facial expression synthesis, and convolutional decoders for joint heatmaps and hand normal maps. These auxiliary branches accelerate convergence and enhance expressive, depth-aware motion representation.

Trained on a diverse collection of human video datasets encompassing a wide range of facial expressions, upper-body, and full-body motions, our method excels at faithfully capturing both intricate and subtle movements and transferring them across subjects with varying identity attributes and poses. We extensively evaluate our model across our challenging motion transfer benchmarks and~\papername~ qualitatively and quantitatively outperforms state-of-the-art human image animation baselines in terms of motion accuracy, identity preservation and visual quality. Moreover, we showcase that our unified identity-disentangled motion tokens facilitate motion and video outpainting beyond human image animation. We summarize our contributions as follows,

\begin{itemize}
\setlength{\itemsep}{3pt}
\setlength{\parskip}{0pt}
\setlength{\parsep}{0pt}
\item A compact, unified, expressive, and depth-aware latent representation for whole-body human motion that is disentangled from identity, capturing complex body and hand articulations, and fine-grained facial expressions.
\item An end-to-end framework that jointly trains a motion image encoder, an identity-aware motion decoder and a DiT-based video generator, with dual decoder guidance to promote semantic and depth-aware motion expressiveness.
\item State-of-the-art performance for cross-identity motion transfer, excelling in motion fidelity, expressiveness, and identity preservation, while also enabling identity-agnostic motion generation and outpainting.
\end{itemize}

\begin{figure*}[ht]
    \centering
    \includegraphics[width=\textwidth]{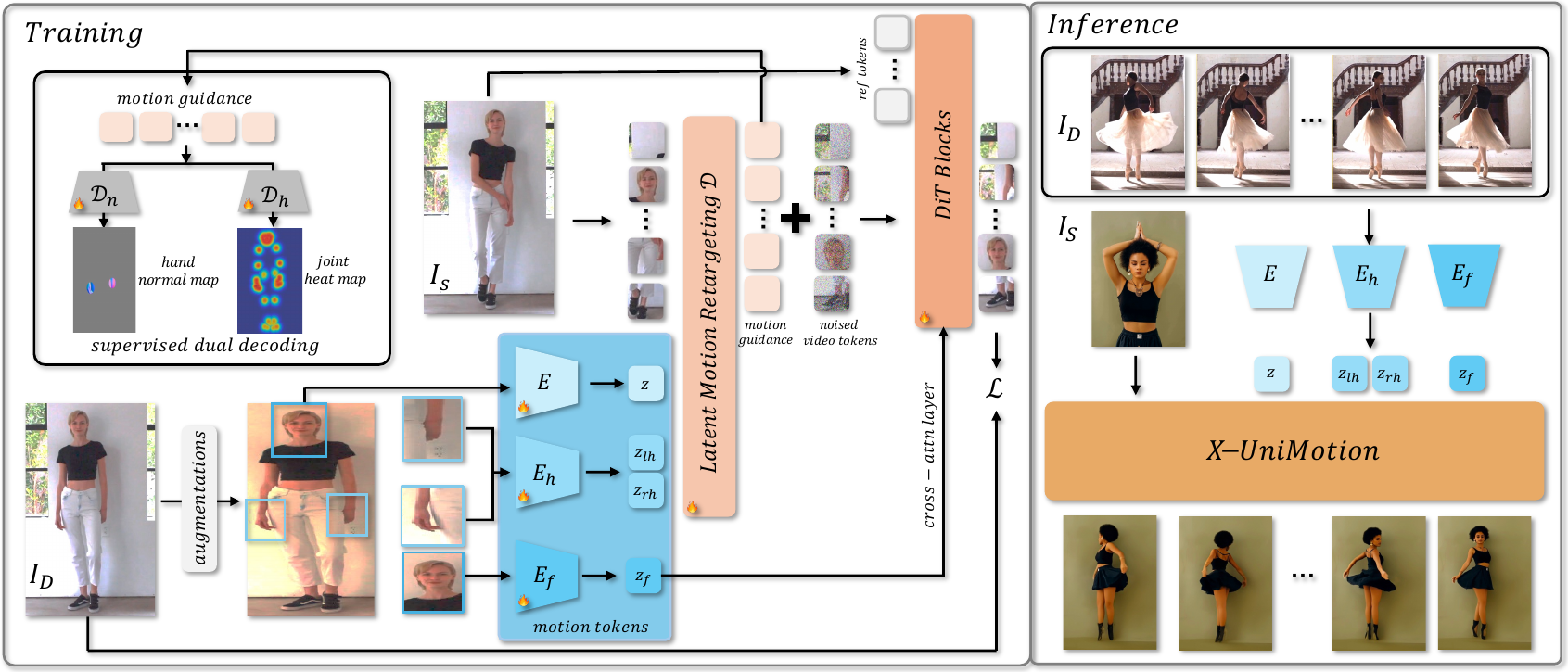}
    \vspace{-6mm}
    \caption{\textbf{Overview.} \papername~features an end-to-end training framework that jointly learns an implicit latent representation of full-body human motion and synthesizes lifelike videos using a DiT network. At its core, we employ an image encoder $\mathcal{E}$ to extract a 1D latent motion descriptor $z$ from the driving image $I_D$, capturing full-body articulation. This global motion code is complemented by decoupled local descriptors—$z_{lh}$ and $z_{rh}$ for the left and right hands, and $z_f$ for facial expressions—extracted from local patches corresponding to the hands and face via $\mathcal{E}_h$ and $\mathcal{E}_f$ respectively. To achieve identity-agnostic motion representation, we apply spatial and color augmentations that effectively disentangle identity cues from the motion latents. These motion tokens are retargeted to the reference subject's body structure in $I_S$ via a ViT decoder $\mathcal{D}$, which outputs identity-aligned spatial motion guidance. This guidance is concatenated with noised video latents and, together with the reference image latents, provided as input to the DiT model. The facial motion latent $z_f$ is further injected into the DiT network via cross-attention layers to control expressions. To supervise motion encoding, we apply dual decoding $\mathcal{D}_h$ and $\mathcal{D}_n$ on the intermediate motion features, predicting both joint heatmaps and hand normal maps. During inference, latent motion codes are directly extracted from each frame of the driving video, enabling expressive and photorealistic animations that maintain strong identity resemblance to the reference image.}
    \label{fig:overview}
    \vspace{-4mm}
\end{figure*}
\section{Related Work}
\noindent\textbf{Diffusion-based Video Generation.}
Diffusion models~\cite{rombach2022high, nichol2021glide, saharia2022photorealistic, ramesh2022hierarchical} have demonstrated strong capabilities in visual generation, with Latent Diffusion Models (LDMs)\cite{rombach2022high} significantly improving efficiency by operating in lower-dimensional latent spaces.
Recent advances in video diffusion models—pretrained on large-scale text-video datasets—have enabled the generation of high-fidelity, temporally coherent videos\cite{kong2024hunyuanvideo, yang2024cogvideox, kuaishoukling, polyak2025moviegencastmedia, wan2025wan}.
In this work, we focus on learning expressive latent motion representations by leveraging a pretrained DiT~\cite{peebles2023scalableDiT} based video diffusion model, i.e., Seaweed-7B~\cite{seawead2025seaweed, lin2025omnihuman}), which serves both to elicit fine-grained motion features and synthesize high-quality human animations.

\noindent\textbf{Human Image Animation.}
Commencing with FOMM~\cite{siarohin2019first}, the task of animating a single human image using motion derived from a driving video—especially across identities—has been extensively studied. Early approaches~\cite{siarohin2021motion,wang2021facevid2vid,zhao2022thin} typically adopt warping and rendering strategies via GANs, where motion is represented with learned 2D or 3D implicit neural keypoints. While effective to some extent, these methods often struggle with modeling extensive movements as well as fine-grained motion details and suffer from limited perceptual quality.
More recent methods leverage large pretrained diffusion models as generative backbones, incorporating various forms of explicit motion representations such as 2D skeletons and facial landmarks~\cite{chang2023magicpose,hu2024animate,zhang2024mimicmotion,wang2025unianimate,lin2025omnihuman,xue2024follow,musepose,wang2024humanvid,xu2024high}, 3D skeletons~\cite{luo2025dreamactor}, dense pose~\cite{xu2024magicanimate}, motion trajectories~\cite{wan2024dragentity}, and 3D parametric models~\cite{zhu2024champ,huang2024make}. MimicMotion~\cite{zhang2024mimicmotion} and HIA~\cite{xu2024high} further model dynamic motion blur with landmark confidence and motion sharpness respectively. While these approaches improve visual fidelity and control, they still fall short in handling nuanced poses and expressions, especially under occlusions or complex articulations. Moreover, the use of explicit motion representations often embeds identity-specific traits (e.g., body shape, facial structure), leading to noticeable identity drifting during cross-identity motion transfer, even when holistic alignments are employed.
To address these limitations, recent works have explored implicit motion modeling. X-Portrait~\cite{xie2024x} introduces patch-based control signals for head pose and expression, while X-NeMo~\cite{xnemo2025} learns identity-agnostic facial motion latents, jointly trained with the generative diffusion model. Animate-X~\cite{tan2024animate} combines both implicit and explicit indicators to enhance motion transfer across anthropomorphic characters.
In contrast, our work advances this direction by introducing a unified, implicit latent representation for expressive whole-body motion—including facial expressions, body pose, and hand gestures. Crucially, our learned motion tokens are disentangled from identity-specific cues, enabling robust and high-fidelity cross-identity motion transfer even under distinct pose variations and identity structures.

\vspace{-2mm}
\section{Method}

Given a single image $I_S$ of a reference subject and a sequence of driving frames $\{I^t_D\}_{t=1}^T$ depicting a different subject performing a motion sequence of $T$ frames, the goal is to transfer human motion across all scales—including full-body articulation, facial expressions, and hand gestures—from the driving frames onto the reference image, producing a temporally coherent and visually faithful animated video.
This task is particularly challenging due to significant discrepancies that may exist between the reference and driving subjects—such as variations in body shape, facial structure, hairstyle and clothing. Moreover, the subjects may appear in different pose configurations (e.g., frontal or crouched), with varying spatial coverage from close-up headshots to full-body frames. These factors compound the difficulty of accurately extracting and retargeting motion while maintaining identity fidelity.

To tackle this, we harness a pretrained Diffusion-Transformer (DiT) model for high-fidelity motion-conditioned video synthesis from the given reference image (Section~\ref{sec:pre}). The core challenge lies in deriving from the driving sequence consecutive motion descriptors that are both expressive and identity-agnostic (Section~\ref{sec:latent}), enabling robust and faithful motion transfer to a novel subject. We propose a self-supervised, end-to-end framework that jointly learns the motion encoder and a motion-conditioned video diffusion model, effectively distilling fine-grained motion cues from rich visual input (Section~\ref{sec:e2e}). We disentangle identity-specific appearance and structure from the learned motion tokens (Section~\ref{sec:id}) and further guide their learning with dual decoders to ensure semantic alignment and fine-grained expressiveness (Section~\ref{sec:dual}).

\subsection{Preliminaries}
\label{sec:pre}
Our video generation framework builds on the Latent Diffusion Model (LDM) paradigm, operating in the compressed latent space of a pretrained causal 3D VAE, which encodes consecutive video frames at adaptive resolutions into compact visual latents.
For the generative backbone, we adopt the MMDiT-based~\cite{esser2024scaling} Seaweed model~\cite{seawead2025seaweed}, extensively pretrained on both text-to-video (T2V) and image-to-video (I2V) tasks. It employs flow matching as the training objective, enabling efficient learning of the reverse denoising process.
To incorporate reference image content, we follow OmniHuman-1~\cite{lin2025omnihuman} by concatenating the flattened latent representation of the source image with the noised video latents. This combined token sequence is fed into the DiT blocks, where self-attention facilitates rich interactions between reference and video tokens, preserving reference identity and background details throughout synthesis.
For motion conditioning, prior approaches typically rely on pose guiders that encode explicit motion cues—such as 2D or 3D skeletal keypoints—into per-frame spatial feature maps. These are stacked with the noise latent along the channel dimension and fed into the diffusion model for structural guidance. While we adopt a similar conditioning mechanism, 
our key innovation is an expressive, learned identity-agonistic motion representation derived directly from image frames, enabling richer dynamics and stronger generalization across identities.

\subsection{End-to-End Learning of Latent Motion Descriptors}
\label{sec:e2e}

As illustrated in Figure~\ref{fig:overview}, we follow an encoder-decoder paradigm, where a motion encoder extracts a latent motion descriptor $z$ from a driving frame $I_D$. This latent is transformed into spatial motion features that guide a DiT-based generative decoder to synthesize a new image $I_{D \rightarrow S}$, in which the reference subject from $I_S$ replicates the motion depicted in $I_D$. The framework is trained in a self-supervised manner as a frame reconstruction task on human motion videos, with $I_S$ and $I_D$ sampled from the same subject. Crucially, we design and train the framework to generalize to cross-identity motion transfer at inference time, where $I_S$ and $I_D$ feature distinct individuals.

Unlike prior approaches that rely on external keypoint detectors to generate explicit skeletal maps from $I_D$—which are then fed into a learnable pose guider—we propose to learn the motion descriptor $z$ implicitly and in an end-to-end fashion. Specifically, we jointly train the motion encoder and the DiT generator, enabling direct extraction of motion embeddings from raw images and seamless conditioning of the generative model. This end-to-end learning fosters mutual reinforcement: the generator incentivizes the encoder to capture semantically rich motion cues, while an informative latent descriptor enhances accuracy and details in synthesized output. Our implicit representation is not limited by the expressiveness of the external motion detectors and can instead capture subtler motion signals, such as pose confidence, motion blur, and local deformations. Moreover, the implicit nature of $z$ eliminates the need for any pose estimation or keypoint preprocessing during inference, resulting in a streamlined and robust motion transfer pipeline that remains effective even under occlusion or extreme lighting.

\subsubsection{Motion Latent Encoding and Retargeting}
\label{sec:latent}

The motion encoder $\mathcal{E}$ takes the driving frame $I_D$ and encodes it into a compact latent motion descriptor $z$ which is identity-agonistic by design. To enable motion retargeting, $z$ is combined with the reference image $I_S$ and passed through an identity-aware decoder $\mathcal{D}$, which transforms the motion latent into a 2D spatial feature map structurally aligned with the identity structure of the reference subject. 

To encode motion, the input image is first resized to a fixed resolution with a square aspect ratio and partitioned into non-overlapping patches of fixed size. The encoder follows a ViT architecture: it linearly projects the image patches and processes them through a stack of transformer blocks to capture long-range dependencies. The resulting feature tokens are then aggregated and passed through a convolutional head to compress them into a single 1D latent vector $z$ that compactly encodes the motion information in $I_D$. Following the information bottleneck principle~\cite{tishby2000information}, we constrain $z$ to be a low-dimensional latent representation (we use 512), serving as a low-pass bottleneck that distills essential motion semantics while discarding high-frequency visual details. Moreover, $z$ is designed as a global, 1D motion descriptor that omits the 2D spatial layout of $I_D$, thereby minimizing the leakage of identity-specific spatial features such as facial structure or body shape. To further regularize the motion latent space and improve generalization in motion generation tasks, we apply a KL divergence loss to $z$. 

\begin{figure*}[t]
\centering
\includegraphics[width=0.95\linewidth]{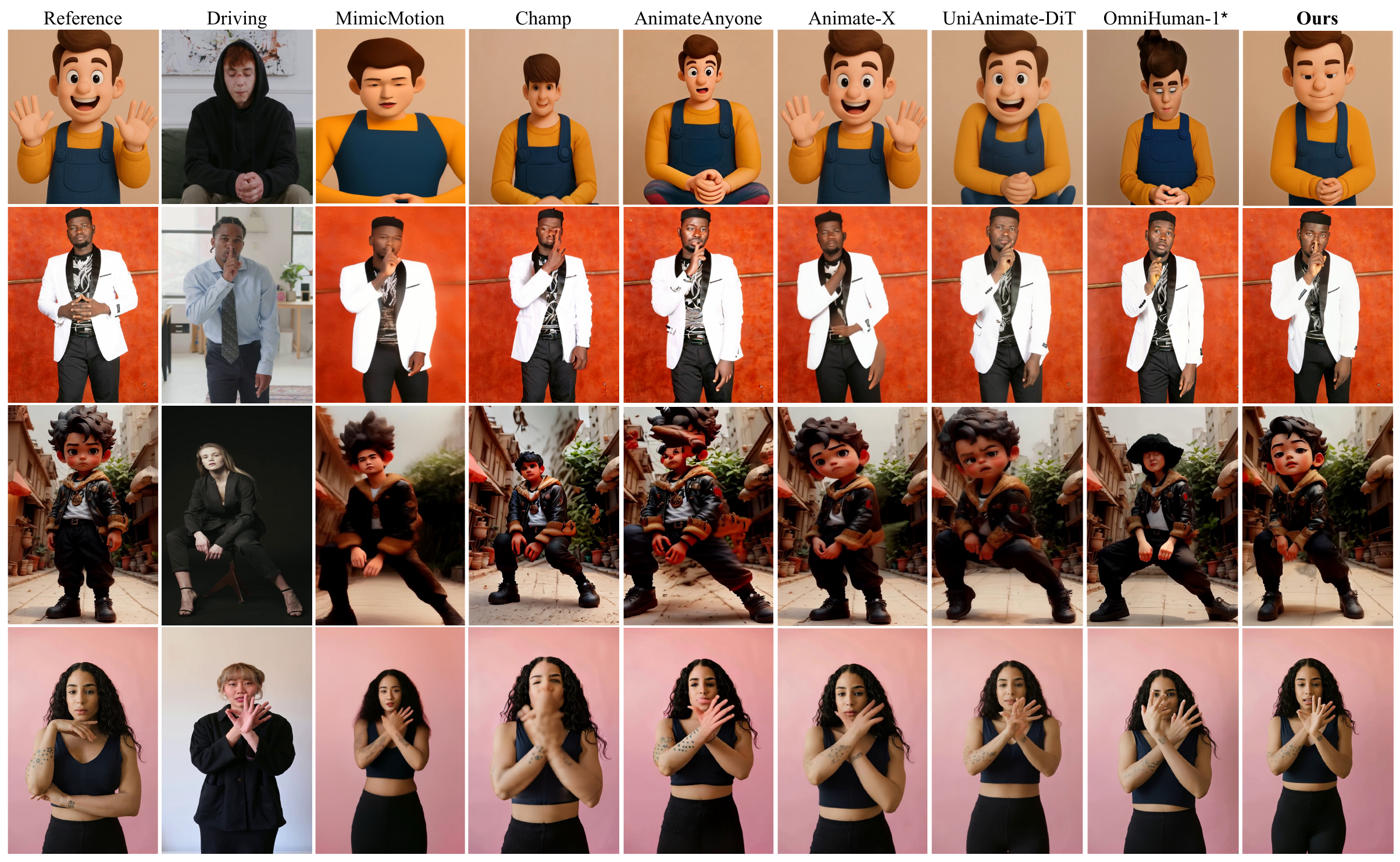}
\vspace{-4mm}
\caption{Qualitative comparison of~\papername~ with SOTA video reenactment baselines. We applied a per-skeleton adjustment on Omni-Human for alignment.}
\vspace{-4mm}
\label{fig:comp}
\end{figure*}

To decode a retargeted spatial motion feature map from $z$ for downstream conditioning in the DiT generator, we employ an asymmetric ViT decoder. This decoder $\mathcal{D}$ takes as input a concatenated token sequence comprising the motion latent $z$ and the linearly projected image patches of $I_S$. Throughout stacked self-attentions, the decoder facilitates deep spatial reasoning and identity-aware motion adaption. From the resulting output tokens, we retain only the updated visual tokens, which embed the motion semantics encoded in $z$ while preserving the structural identity cues of $I_S$.

\subsubsection{Motion-Identity Disentanglement}
\label{sec:id}
A naive end-to-end training of the above pipeline tends to entangle identity information within the motion latent embedding $z$, as the driving and target image are the same in our self-supervised training.  
To promote identity-agonistic motion embedding, we apply 
motion-invariant augmentations to $I_D,$ including color jittering, random scaling within 30\%, and piecewise affine transformations. 
These augmented input pairs introduce contrastive identity cues, which incentivize the motion encoder to distill motion signals exclusively from the driving frame, while guiding the motion decoder and DiT generator to rely on $I_S$ for appearance and structural identity features. 

Nevertheless, such 2D image augmentations do not alter part-wise proportions, and identity leakage remains observable—particularly in scenarios where the reference subject exhibits atypical body ratios, such as 3D characters with disproportionately large heads or hands. To further enhance identity disentanglement in motion latent learning, we introduce synthetic training pairs by rendering randomly selected 3D characters under identical body poses. We apply random scaling on individual skeletal limbs as well as to the sizes of head and hands, effectively simulating challenging heterogeneous cross-identity motion transfer and encouraging the model to decouple motion from identity structural cues. Notably the 2D augmented images remain essential, as they provide diverse and detailed human motions from widely accessible human videos. 

\subsubsection {Localized Motion Descriptors}
\label{sec:local}
To this end, our latent motion descriptor is designed as a single embedding aimed at encoding full-scale human motion from $I_D$. However, encoding fine-grained movements—such as finger articulation and subtle facial expressions—within a compact representation is inherently challenging. Increasing the dimensionality of the latent may improve expressiveness, but it also increases the risk of entangling high-frequency visual features, undermining motion-identity disentanglement.

To address this, we introduce three additional localized motion descriptors: $z_f$ for the face, and $z_{lh}$, $z_{rh}$ for the left and right hands, respectively. Together with the global descriptor $z$, they form the complete motion embedding $z_{\text{full}} = (z, z_f, z_{lh}, z_{rh})$, where each component shares the same dimensionality.  Each local descriptor focuses on capturing fine-grained motion cues within its respective cropped region.
To obtain these local embeddings, we employ two additional motion encoders $\mathcal{E}_{f}, \mathcal{E}_{h}$ with the same ViT architecture as $\mathcal{E}$. These encoders operate on square crops of the face and hands from $I_D$, with a shared encoder $\mathcal{E}_h$ for both hands by horizontally flipping the right-hand patch. The hand descriptors $z_{lh}$ and $z_{rh}$, along with the global $z$ and patches of $I_S$, are jointly processed by our ViT decoder $\mathcal{D}$ to synthesize the retargeted spatial motion features.  Notably, following~\cite{xnemo2025}, we integrate the facial descriptor $z_f$ into the DiT generator using a cross-attention mechanism rather than a pose guider, facilitating better facial identity disentanglement. Although $z_f$, $z_{lh}$, and $z_{rh}$ are local descriptors, both the ViT decoder $\mathcal{D}$ and the DiT generator learn to associate them with their respective spatial regions through attentions.


\begin{figure*}[t]
\centering
\includegraphics[width=1.0\linewidth]{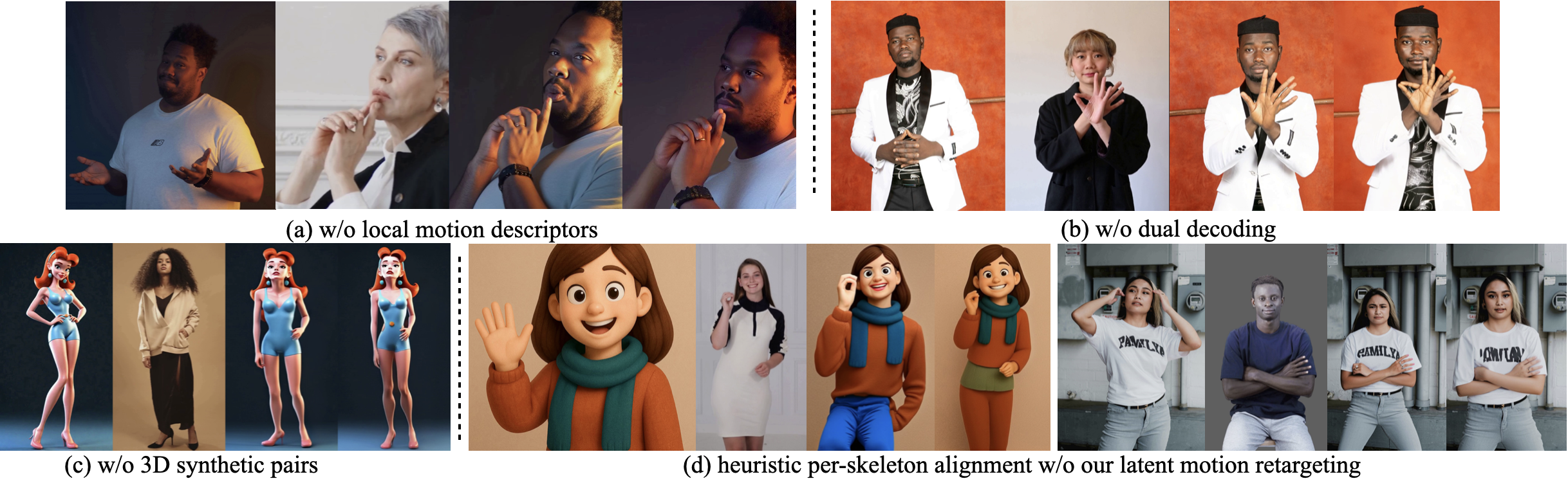}
\vspace{-8mm}
\caption{Qualitative Ablation Study. From left to right: reference image, driving image, result from ablated model, and our full method.  }
\vspace{-4mm}
\label{fig:abla}
\end{figure*}

\vspace{-2mm}
\subsubsection{Supervised Dual Decoding}
\label{sec:dual}
In our early experiments, we observe that a straightforward end-to-end training of the full pipeline often converges slowly and fails to effectively capture intricate yet identity-agnostic human motions in the latent descriptors $z_{\text{full}}$. One key limitation arises from the diffusion loss, which assigns equal weight to all pixels—without explicitly prioritizing the modeling of structured human motion over scene-specific or identity-specific dynamics such as background, hair, and clothing.
Furthermore, the powerful DiT backbone may learn to hallucinate plausible motion dynamics using its internal temporal priors, thereby reducing reliance on the encoded motion latent $z_{\text{full}}$ and undermining the informativeness of the latent motion representation. 

To address the limitations of diffusion-only training, we introduce dual motion decoders equipped with supervised motion cues to guide the learning of the latent motion embedding toward the desired granularity and semantics of human motion. Specifically, we attach two lightweight convolutional decoder heads, $\mathcal{D}_h$ and $\mathcal{D}_n$, which operate on the spatial motion features produced by the ViT decoder $\mathcal{D}$. These heads are trained to regress per-joint heatmaps and hand normal maps extracted using Sapiens~\cite{khirodkar2024sapiens}, thus providing explicit supervisory signals for both global skeletal pose and fine-grained hand articulation.
Notably, the supervision from hand normal maps effectively resolves inherent ambiguities in 2D skeletons—such as depth ordering and projection ambiguity—significantly improving the model's ability to handle complex hand gestures and inter-hand interactions. We avoid using full-body depth maps for supervision, as they tend to encode undesirable identity-specific features (e.g., clothing and hairstyles). For latent facial motion supervision, we follow the approach of~\cite{xnemo2025}, employing a dual GAN-based renderer~\cite{wang2022pdfgc}.


\begin{table*}[t!]
\centering
\caption{Quantitative comparisons of \papername~ with SOTA video reenactment baselines. *We apply per-skeleton bone-length alignment on OmniHuman~\cite{lin2025omnihuman}.}
\vspace{-3mm}
\scalebox{1}{
\begin{tabular}{lccccc|ccc}
\toprule
\multirow{2}{*}{\textbf{Methods}} 
& \multicolumn{5}{c|}{\textbf{Self Reenactment}} 
& \multicolumn{3}{c}{\textbf{Cross Reenactment}} \\
\cmidrule(lr){2-6} \cmidrule(lr){7-9}

& \textbf{SSIM} $\uparrow$  
& \textbf{PSNR} $\uparrow$ 
& \textbf{LPIPS} $\downarrow$  
& \textbf{FID} $\downarrow$ 
& \textbf{FVD} $\downarrow$ 
& \textbf{FaceID-Sim} $\uparrow$ 
& \textbf{FullID-Sim} $\uparrow$
& \textbf{Mot-Acc} $\uparrow$
\\
\midrule
MimicMotion~\cite{zhang2024mimicmotion}     & 0.67 & 16.50 & 0.35 & 90.82 & 965.58 & 0.27 & 2\%  & 4\% \\
Champ~\cite{zhu2024champ}                   & 0.69 & 17.14 & 0.32 & 61.56 & 921.42 & 0.22 & 1.3\%  & 0\% \\
AnimateAnyone~\cite{hu2024animate}          & 0.70 & 18.81 & 0.29 & 58.20 & 592.61 & 0.42 & 2.3\%  & 1.3\% \\
Animate-X~\cite{AnimateX2025}               & 0.79 & 22.18 & 0.23 & 48.78 & 380.37 & 0.56 & 15\%  & 3\% \\
UniAnimate-DiT~\cite{wang2025unianimate}    & 0.80 & 22.71 & 0.21 & 41.95 & 235.22 & 0.54 & 13.7\%  & 11.7\% \\
OmniHuman$^{*}$~\cite{lin2025omnihuman} & 0.74 & 20.85 & 0.23 & 39.86 & 291.28 & 0.51 & 10\% & 9.7\% \\
\midrule
\textbf{X-UniMotion (Ours)}   & \textbf{0.83} & \textbf{23.81} & \textbf{0.19} & \textbf{36.91} & \textbf{176.45} & \textbf{0.61} & \textbf{55.7}\% & \textbf{70.3}\% \\
\bottomrule
\end{tabular}}
\vspace{-3mm}
\label{tab:comp}
\end{table*}

\section{Experiments}
\subsection{Implementation Details}
\label{sec:dataset}
\noindent\textbf{Datasets.} We construct a training set comprising approximately 200 hours of in-house video footage, capturing a broad range of real human body motions, hand gestures, and facial expressions. To enhance cross-identity generalization, we supplement this with a 3-hour synthetic dataset featuring 500 3D-rigged human characters animated with diverse motion sequences from Mixamo~\cite{mixamo}. The overall dataset is balanced between full-body and half-body shots, each constituting roughly half of the content.

We compile an evaluation benchmark of 100 in-the-wild reference human images  collected from Midjourney~\cite{midjourney} and Pexels~\cite{pexels}, encompassing diverse body and facial structures (e.g., adults, children, cartoonish, humanoid), styles (e.g., oil painting, 3D cartoon), and spatial compositions (e.g., full-body, upper-body, close-up). In addition, we gather 100 in-the-wild driving clips that feature challenging body motions (e.g., yoga, crouching), intricate intra-hand gestures, and expressive facial dynamics. Please refer to our supplementary webpage for additional video results.

\noindent\textbf{Training and Inference.} 
We utilize a pre-trained DiT model\cite{seawead2025seaweed}, along with its appearance conditioning module~\cite{lin2025omnihuman}, as the generative backbone of our framework. On our curated training dataset, we fine-tune the DiT module using the AdamW optimizer with a learning rate of $1e^{-6}$, while jointly training the latent motion encoding and retargeting modules with a learning rate of $1e^{-4}$. Training is conducted across 80 A100 GPUs for 40K steps, using video clips of varying durations ranging from 53 to 121 frames, preserving the original aspect ratio.
During inference, video generation is performed in chunks of 75 frames using a sliding window with a 4-frame overlap to ensure temporal consistency. We apply classifier-free guidance (CFG) with a guidance scale of 2 for both the reference appearance and motion control signals.

\vspace{-2mm}
\subsection{Evaluations and Comparisons}
We evaluate our method against state-of-the-art video-driven human image animation baselines, including SD UNet-based approaches such as MimicMotion~\cite{zhang2024mimicmotion}, Champ~\cite{zhu2024champ}, AnimateAnyone~\cite{hu2024animate}, and Animate-X~\cite{AnimateX2025}, as well as the DiT-based UniAnimate-DiT~\cite{wang2025unianimate} built on Wan2.1-14B~\cite{wan2.1} and OmniHuman-1~\cite{lin2025omnihuman} based on Seaweed-7B~\cite{seawead2025seaweed}. We assess performance under both self-and cross-identity reenactment settings. All evaluation metrics are computed on outputs with a resolution of $720 \times 1280$.


\noindent\textbf{Self Reenactment.}  For each test video, we use the first frame as the reference image and generate the full sequence, where the remaining frames serve as both the driving signal and the ground-truth targets.
We evaluate the motion reenactment performance using a combination of pixel-wise (PSNR$\uparrow$), structural (SSIM$\uparrow$), and perceptual (LPIPS$\downarrow$) image similarity metrics. To assess overall video quality and motion realism, we also report FID$\downarrow$ and FVD$\downarrow$.
As shown in Table~\ref{tab:comp}, \papername~consistently outperforms all baseline methods across all evaluation metrics.

\begin{figure*}
\centering
\includegraphics[width=0.98\linewidth]{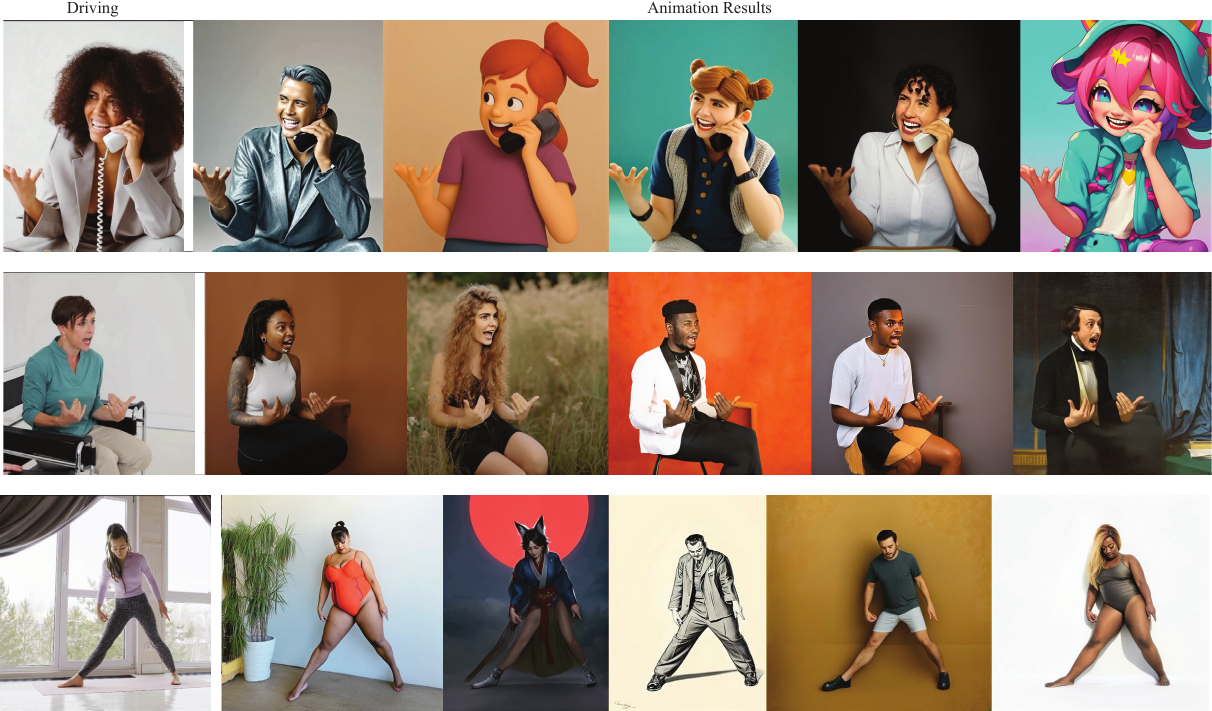}
\vspace{-4mm}
\caption{More qualitative results.The leftmost column shows the driving frame, and we show the animated results for multiple images by \papername.  }
\vspace{-4mm}
\label{fig:big0}
\end{figure*}

\begin{figure*}[t]
\centering
\includegraphics[width=0.99\linewidth]{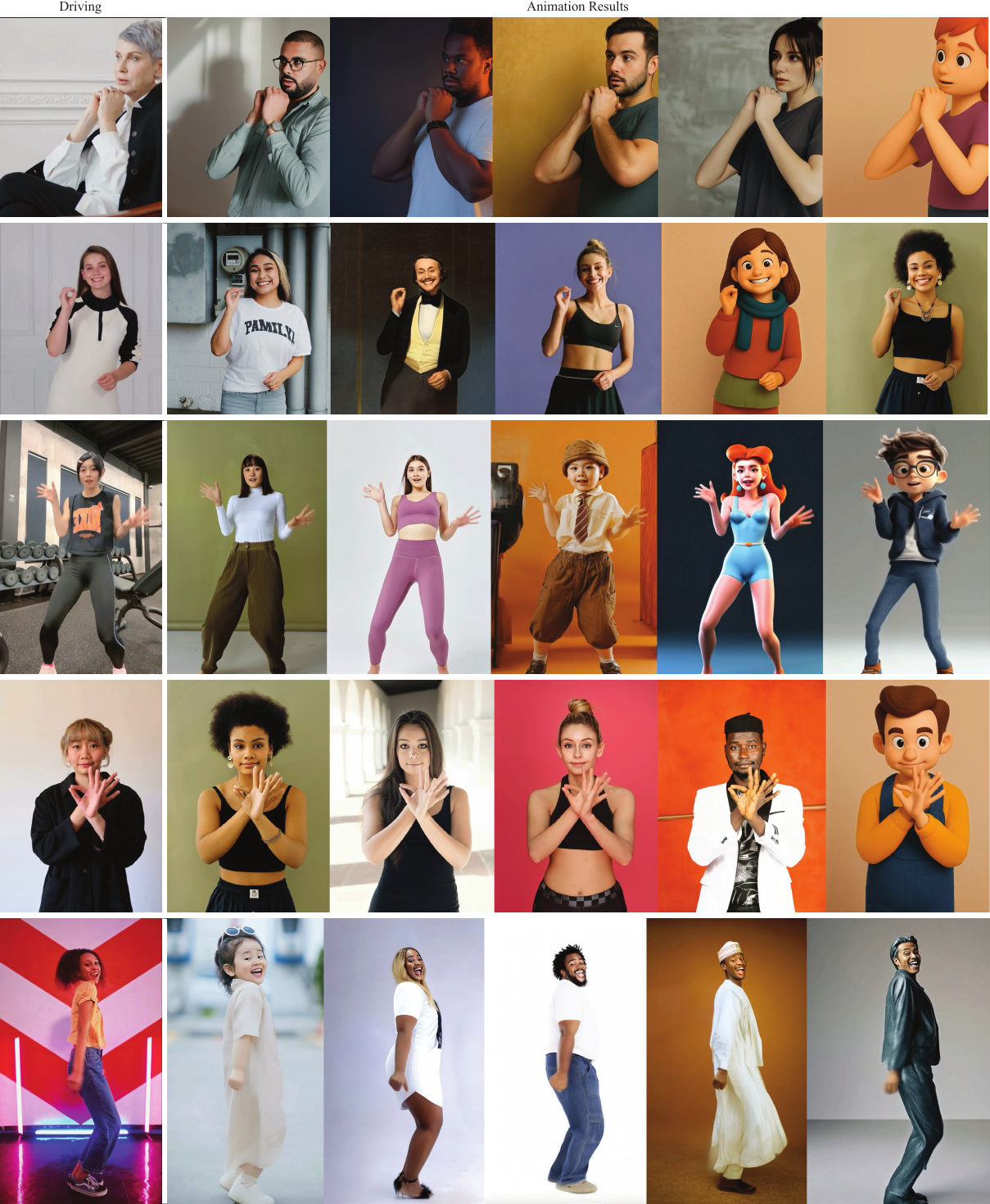}
\caption{More qualitative results.The leftmost column shows the driving frame, and we show the animated results for multiple images by \papername.}
\label{fig:big1}
\end{figure*}

\noindent\textbf{Cross Reenactment}.
Our method enables the generation of captivating and expressive animations across a wide range of human images, even when driven by in-the-wild videos with varying identity traits, poses, and spatial configurations (Figure~\ref{fig:teaser},~\ref{fig:comp},~\ref{fig:big0},~\ref{fig:big1}). As shown in our qualitative comparisons (Figure~\ref{fig:comp}), \papername~ significantly outperforms all baseline methods in terms of identity preservation, motion accuracy, and perceptual quality.
Notably, all 2D skeleton-based methods—namely MimicMotion, AnimateAnyone, UniAnimate-DiT and OmniHuman-1 are evaluated with per-skeleton alignment. However, due to the inherent 2D projection ambiguities, these methods struggle to faithfully adapt to the reference subject’s body structure, especially when the driving video differs significantly in viewpoint (e.g., frontal vs. side profile), pose (e.g., sitting vs. standing), or spatial composition (e.g., half-body vs. full-body). 
These keypoint-based approaches also lack the expressiveness needed for fine-grained motions, such as eye squinting (1st and 3rd rows), lip puckering (2nd row), finger pointing (2nd row), and pinching gestures (4th row). Furthermore, they are prone to depth ambiguity and occlusion issues, as seen in crossing hands and fingers (1st and 4th rows).
Similarly, Champ, which employs 3D SMPL-X~\cite{SMPL-X:2019}, and Animate-X, which integrates both explicit and implicit pose indicators, offer limited expressiveness for complex motions. 
Animate-X in particular, though designed for heterogeneous body structures, often defaults to static poses that fail to follow the driving motion (1st and 2nd row).
In contrast, \papername~leverages a unified, identity-agnostic motion representation that captures multi-scale human motion—ranging from subtle facial dynamics to full-body articulation—while preserving the identity of the reference subject with high fidelity.


For quantitative evaluation, we adopt ID-SIM$\downarrow$ (computed with ArcFace~\cite{deng2019arcface}) to assess face-region identity similarity (FaceID-Sim$\uparrow$) in the absence of ground-truth videos. We further conduct a user study with 60 participants, each comparing 10 randomly selected pairs, to evaluate full-body identity similarity (FullID-Sim$\uparrow$) and motion expressiveness (Mot-Acc$\uparrow$). As shown in Table~\ref{tab:comp}, our method outperforms all baselines across all metrics.


\vspace{-2mm}
\subsection{Ablation Studies}


We ablate key components of our framework by removing them from the full pipeline, evaluated on cross-identity reenactments. 
Figure~\ref{fig:abla}(a) highlights the importance of local motion descriptors in capturing fine-grained facial expressions and hand dynamics. When relying solely on a single global motion descriptor, we observe misalignments in eye gaze, lip shapes, finger positions and their interleaving status.
We also evaluate the contribution of our dual decoding mechanism in Figure~\ref{fig:abla}(b). Omitting this module compromises both motion detail and depth understanding, leading to incorrect hand gestures and reversed hand ordering.
Figure~\ref{fig:abla}(c) examines the impact of 3D synthetic cross-identity rendering pairs, which play a crucial role in promoting identity-motion disentanglement—particularly for characters with non-human-like body proportions.
Finally, Figure~\ref{fig:abla}(d) shows the effect of replacing our motion representation and retargeting module with explicit 2D skeleton conditioning on the DiT video backbone. Even under optimal per-skeleton alignment, the system fails to preserve identity traits—resulting in inaccurate body, hand, and head proportions—and struggles with depth ordering and occlusion handling.

To quantitatively assess the efficacy of each ablated module, we evaluate performance on our test set of synthetic cross-identity 3D pairs. Specifically, we report the L1 distance between normalized ground-truth and detected full-body and hand keypoints (Table~\ref{tab:abla}). We omit image-space metrics, as the synthetic renderings are not photorealistic; instead, landmark differences effectively capture both motion expressiveness and identity adaptability.
As shown in Table~\ref{tab:abla}, both the local motion descriptors and dual decoding significantly boost motion expressiveness, especially in local regions. The use of 3D synthetic pairs enhances identity-awareness in motion retargeting across diverse characters. In contrast, explicit per-skeleton alignment yields the weakest results, underscoring the necessity of our proposed latent motion embedding and retargeting strategy.

\subsection{Applications}


Our latent motion descriptor serves as a unified representation for both motion comprehension and generation, enabling applications beyond video retargeting—such as video outpainting via continuous motion prediction. Specifically, we train a motion diffusion model to predict 81 future frames of latent motion codes, conditioned on the preceding 9 frames. As illustrated in Figure~\ref{fig:app}, our latent representation facilitates the generation of natural and expressive full-body motion sequences, which can be seamlessly retargeted to different subjects owing to its identity-disentangled nature.

\begin{table}
\caption{Quantitative ablation study, reporting mean L1 differences on normalized key points of full body (KP) and local hands (KP-H).}
\vspace{-3mm}
    \centering
    \begin{tabular}{l|ccc}
    \hline
    \textbf{Methods}                           &  \textbf{KP} $\downarrow$ & \textbf{KP-H} $\downarrow$  \\ \hline
    (a) w/o local motion descriptors                 & 0.070&0.077 \\
    (b) w/o dual decoding                          & 0.057&0.058 \\
    (c) w/o 3D synthetic pairs                      & 0.049 &0.057  \\
    (d) heuristic per-skeleton alignment        & 0.095 &0.114\\
    \midrule
    \textbf{\papername~(Ours)}                             &  \textbf{0.046} & \textbf{0.052}\\ \hline
    \end{tabular}
\vspace{-3mm}
\label{tab:abla}
\end{table}


\section{Conclusion}
We introduced \papername, a unified, identity-agnostic motion representation that enables expressive and high-fidelity human image animation across subjects. By leveraging a self-supervised training framework with joint motion encoding and DiT-based video generation, along with auxiliary spatial decoders, our method captures fine-grained, depth-aware motions, spanning full-body articulations, subtle facial expressions, and intricate hand gestures. Extensive evaluations demonstrate state-of-the-art performance in motion fidelity, identity preservation, and generalization to challenging cross-identity motion transfer with distinct identities and poses.

\begin{figure*}[ht]
\centering

\begin{minipage}[t]{0.48\linewidth}in
    \centering
    \includegraphics[width=0.9\linewidth]{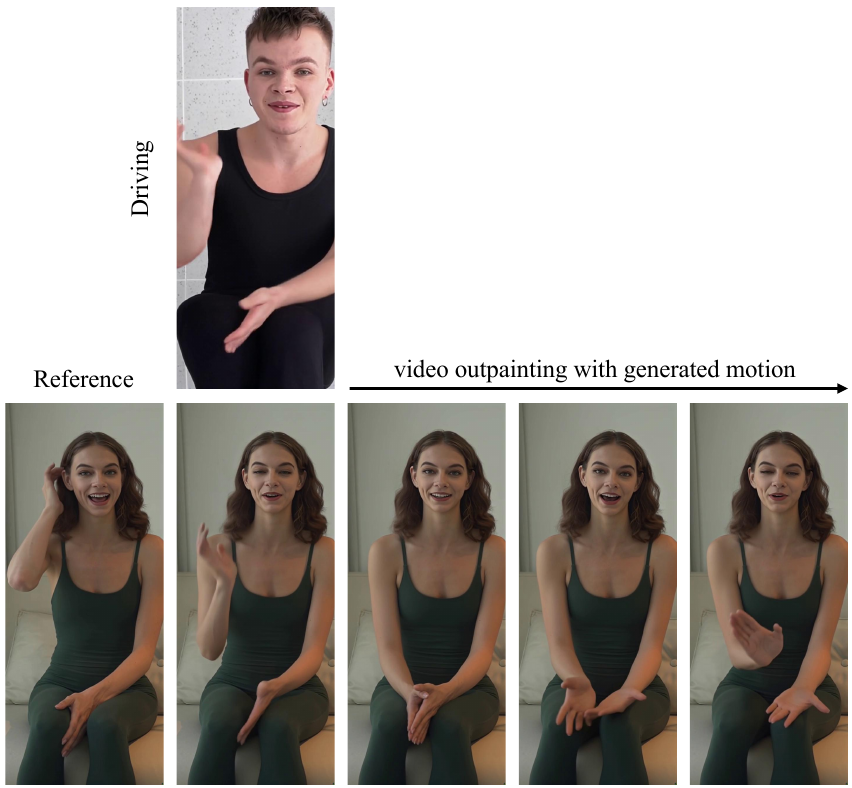}
    \caption{Video reenactment and outpainting with extrapolated latent motion generation. Our model extracts motion from a short input sequence and retargets it to a different subject, extrapolating into a longer video that maintains coherent identity and generates natural, expressive full-body motion—including detailed facial expressions and hand gestures.}
    \label{fig:app}
\end{minipage}
\hfill
\begin{minipage}[t]{0.48\linewidth}
    \centering
    \includegraphics[width=\linewidth]{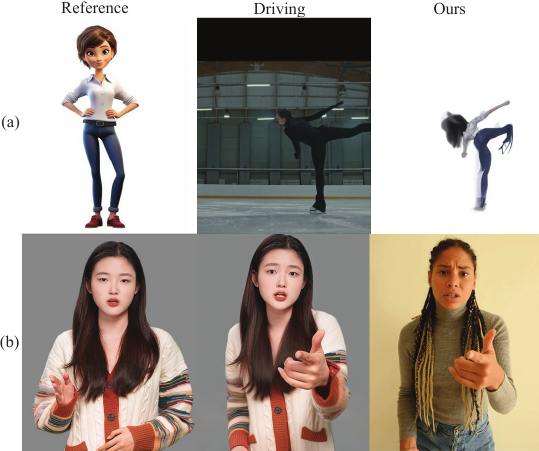}
    \caption{Limitations. \papername~may struggle with highly challenging out-of-domain motions and can occasionally miss subtle expressions such as frowning.}
    \label{fig:limitation}
\end{minipage}
\end{figure*}

\noindent\textbf{Limitations and Future Work.} 
While \papername~demonstrates strong performance in single-person human image animation, it currently does not address multi-person scenarios or interactions. In future work, we aim to extend our framework to model human-human and human-object interactions. Additionally, our method is designed for humans or anthropomorphic characters, and adapting it to heterogeneous subjects such as animals presents an exciting direction. Lastly, the expressiveness of our latent motion representation could be further enhanced by training on more diverse and challenging video datasets, as shown in Figure~\ref{fig:limitation}.

\bibliographystyle{ACM-Reference-Format}
\bibliography{main}

\clearpage 

\clearpage 




\end{document}